\documentclass[runningheads]{llncs}
\usepackage[T1]{fontenc}
\usepackage{amsmath}
\usepackage{amsfonts}
\usepackage{algorithm}
\usepackage{xcolor}
\usepackage{pgfplots}

\usepackage{graphicx}
\usepackage{wrapfig}

\usepgfplotslibrary{groupplots}
\pgfplotsset{compat=1.18}
\begin{document}
\title{Similarity-Aware Machine Unlearning}
%
%
\author{Madhavan Citalamangalam Kumaran\inst{1}\orcidID{0009-0008-9954-090X} \and
Midhun Parakkal Unni\inst{2,3}\orcidID{0000-0002-2913-8869}\and
Vicky Kouni\inst{4}\orcidID{0000-0002-1316-8967}\and
Haripriya Harikumar\inst{1}\orcidID{0000-0001-9918-381X}}
\authorrunning{Madhavan Citalamangalam Kumaran et al.}
%
\institute{Department of Computer Science, University of Manchester, UK \and
Centre for Machine Intelligence, University of Sheffield, UK\and
School of Computer Science, University of Sheffield, Sheffield, UK\and
Paris Dauphine - PSL University, Paris, France\\
\email{madhavan.citalamangalamkumaran@student.manchester.ac.uk,
m.parakkalunni@sheffield.ac.uk,
vasiliki.kouni@lamsade.dauphine.fr,
haripriya.harikumar@manchester.ac.uk
}
}
\maketitle              
\begin{abstract}
Machine unlearning removes the influence of user-specified training examples from a trained model, avoiding the need to retrain it from scratch. Localization-based methods improve unlearning efficiency by identifying a subset of influential model parameters. However, existing approaches select parameters based solely on forget-set importance, neglecting their role in retained dataset and often causing collateral damage to semantically similar retained examples. We address this limitation with a retain-aware localization method that considers parameter importance to both forgotten and retained data. We also introduce a retain-similar evaluation set, constructed using cosine similarity in the model embedding space, to directly measure collateral damage. Across eleven experimental settings on CIFAR-10 dataset and ResNet18 model, our method consistently reduces collateral damage while improving standard unlearning metrics, demonstrating the effectiveness of retain-aware localization for similarity-aware machine unlearning.
\keywords{Similarity  \and Unlearning \and collateral damage \and embedding space \and localization}
\end{abstract}
\section{Introduction}
Machine learning models can memorize information about individual training examples to varying degrees \cite{wei2025memorization}, creating privacy risks even when the underlying training records are not directly accessible. Membership-inference attacks can reveal whether an example was used for training \cite{shokri2017membership}, model-inversion attacks \cite{fredrikson2015model} can expose sensitive attributes \cite{schwartz2004property}, and generative models can reproduce passages from their training corpora verbatim \cite{carlini2021extracting}. These risks are amplified when models are trained on personal or sensitive data \cite{raab1998distribution,schwartz2004property}. Data-protection frameworks such as the General Data Protection Regulation \cite{voigt2017eu} and California Consumer Privacy Act \cite{pardau2018california} establish rights concerning the erasure or deletion of personal data. Together, these privacy risks and data-deletion rights motivate mechanisms for removing the residual influence from models. 

Machine unlearning \cite{cao2015towards,bourtoule2020machineunlearning,xu2024machine} addresses this problem by modifying a trained model so that its behavior approximates that of an oracle model retrained without a user-specified subset of training examples, referred to as the forget set. Retraining from scratch on the remaining data provides the best solution, but can be computationally expensive, particularly when deletion requests are frequent. Approximate unlearning \cite{nguyen2020Varational} methods therefore seek to approximate the behavior of the retrained model at substantially lower computational cost. Localized unlearning \cite{torkzadehmahani2024improved} methods reduce this cost by restricting model modifications to a selected subset of parameters that are influential for the forget set.

\begin{wrapfigure}{r}{0.5\textwidth}
    \centering
    \includegraphics[width=0.48\textwidth]{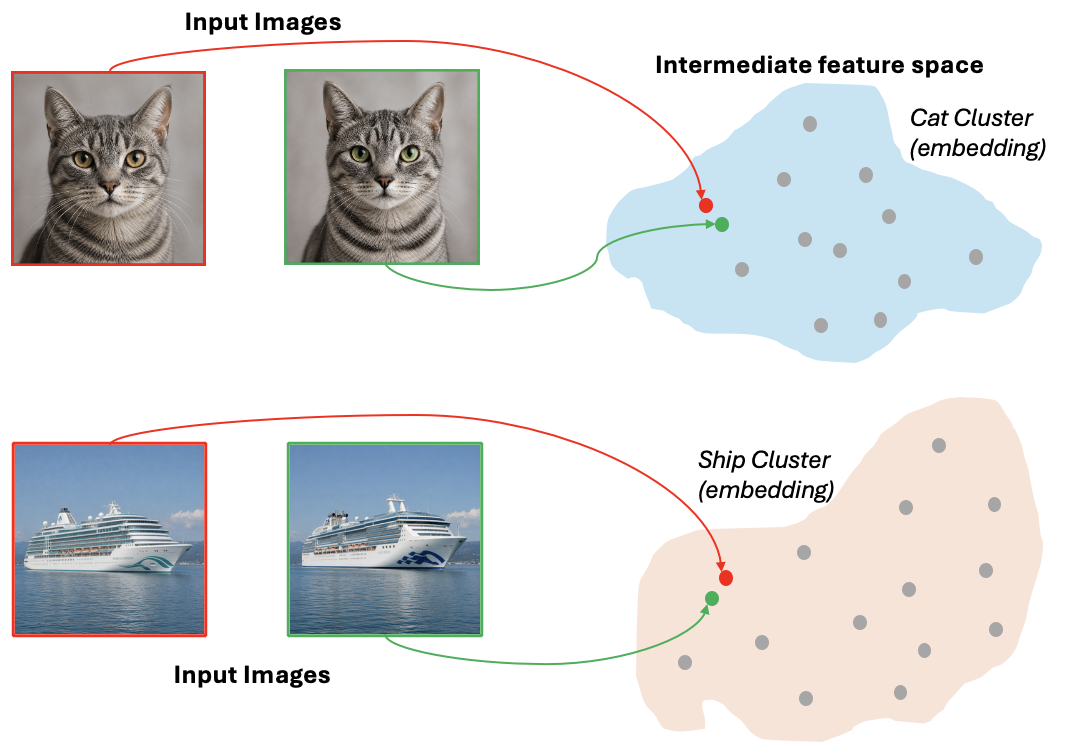}
    \caption{Intermediate feature-space visualization showing that semantically similar samples are embedded close to one another within class-specific clusters. The image to forget (red) and the image to retain (green) occupy very close neighboring regions of the feature manifold, indicating shared feature representations.}
    \label{fig:intro_similar}
\end{wrapfigure}

However, importance to the forget set does not imply specificity to the forget set. A parameter may be highly influential for forgotten examples while also supporting retained examples. This overlap is especially consequential when forgotten and retained examples occupy nearby regions of the learned representation space and may depend on shared features, as shown in Fig. \ref{fig:intro_similar} (visually similar images in red and green boxes naturally cluster closely in their embedding space).Consequently, forget-only localization can modify parameters used by both sets, causing unintended degradation on representation-similar retained examples. We refer to this representational overlap as \textit{feature entanglement} and to the resulting degradation as \textit{collateral damage}. Because such damage may be concentrated among a small subset of retained examples, aggregate test and retain-set accuracy may fail to reveal it. 

To address this limitation, we propose a retain-aware localization framework. We compute criticality scores separately on the forget set and the retain set, and combine them to prioritize neurons that are important to the forget set but comparatively less important to the retain set. To measure collateral
damage directly, we also construct a retain-similar evaluation set. For each forgotten example, this set contains its nearest same-class retained neighbor according to cosine similarity in the original model's embedding space.
Our contributions are as follows.
\begin{itemize}
    \item We identify collateral damage to semantically similar retained examples as a key limitation of existing localization-based machine unlearning methods and relate this phenomenon to shared learned representations.
    \item We propose a retain-aware localization framework that incorporates parameter importance with respect to both forgotten and retained data, enabling similarity-aware unlearning.
    \item We introduce a retain-similar evaluation protocol and demonstrate across eleven experimental settings that our approach consistently reduces collateral damage while improving standard unlearning metrics.
\end{itemize}

\section{Background}
We introduce the notation and definitions used throughout the paper before presenting our method. Let $\mathcal{D} := \bigcup_{n \geq 1} (\mathcal{X} \times \mathcal{Y})^n$ denote the set of all datasets and $\mathcal{A}$ a randomized training algorithm,$
\mathcal{A} : \mathcal{D} \rightarrow \Delta(\mathcal{H})$,
where $\Delta(\mathcal{H})$ denotes the set of all probability distributions over the hypothesis space $\mathcal{H}$. The hypothesis space $\mathcal{H}:=\{f_\theta:\mathcal{X}\to\mathcal{Y} \mid \theta\in \mathbb{R}^d\}$ is the set of all models that the learning algorithm $\mathcal{A}$ produces, parameterized by weight vectors $\theta \in \mathbb{R}^d$.  We denote by $f_{\theta^o} \sim \mathcal{A}(D_{train}\subset \mathcal{D})$ the original model parameterised by $\theta^o$ when trained on the training set $D_{train}\subset \mathcal{D}$ prior to unlearning.

Let the forget set be denoted by $D_f \subset D_{train}$, consisting of the data requested for removal. The retain set is defined as,$D_r = D_{train} \setminus D_f$ representing the remaining training data. We denote $f_{\theta^r} \sim \mathcal{A}(D_r)$ as oracle model and $\theta^r$ as the associated parameters by training from scratch only on the retain set alone. Machine unlearning is defined as the process of modifying the  parameters of a model from $\mathcal{A}(D_{train})$, in order to remove the influence of $D_f$ without compromising performance on the retain set $D_r$. Formally,
\begin{definition}[Unlearning \cite{bourtoule2021machine}]
\label{unlearningdef}
An unlearning algorithm $\mathcal{U}$ is said to unlearn a forget set $D_f \subset D_{\text{train}}$ if the unlearned model $f_{\theta^u} \sim \mathcal{U}(f_{\theta^o}, D_f, D_r)$ is indistinguishable from the retrained model $f_{\theta^r} \sim \mathcal{A}(D_r)$, where $D_r = D_{\text{train}} \setminus D_f$. Formally, the output distributions of $f_{\theta^u}$ and $f_{\theta^r}$ are identical on the evaluation distribution.
\end{definition}
Definition 1 characterizes the goal of machine unlearning where the unlearned model $f_{\theta^u}$
should behave indistinguishably from a model retrained from scratch on the retain set $D_r$. One way to satisfy this definition is exact unlearning, where the model is retrained from scratch using only $D_r$, to produce $f_{\theta^r} \sim \mathcal{A}(D_r)$. Although exact unlearning directly satisfies the definition, retraining with large models and datasets can be computationally expensive. Therefore, practical approaches often employ approximate unlearning, where an unlearning algorithm $\mathcal{U}$ modifies the original model $f_{\theta^o}$ to obtain $f_{\theta^u} \sim \mathcal{U}(f_{\theta^o},D_f,D_r)$. The objective is to have an unlearned model $f_{\theta^u}$ that produces outputs closely matched to those of the retrained model $f_{\theta^r}$, while requiring significantly less computation.
More precisely, machine unlearning approach aims for the following:
\begin{align}
    t\bigl(\mathcal{U}(f_{\theta^o},\, D_f, D_r)\bigr) 
        &\ll t\bigl(\mathcal{A}(D_r)\bigr), \\[6pt]
    \mathcal{U}(f_{\theta^o},\, D_f, D_r) 
        &\approx \mathcal{A}(D_r),
\end{align}
where $t(\cdot)$ denotes the execution time of the unlearning or training process. We use the following definition to operationalise evaluation of unlearning in line with Torkzadehmahani et al. \cite{torkzadehmahani2024improved}.
\begin{definition}[Localized Machine Unlearning \cite{torkzadehmahani2024improved}]
Let $\theta^o = \{\theta_k\}_{k=1}^{N}$ denote the parameters of the original model, where $N$ is the total number of trainable parameters. An unlearning algorithm $\mathcal{U}$ is said to perform localized machine unlearning if, given a forget set $D_f$, it identifies and modifies only a subset of parameters $\theta_s \subset \theta^o, |\theta_s|=n< N,$ while leaving the remaining parameters unchanged during unlearning. The resulting unlearned model is from $\mathcal{U}(f_{\theta^o}, D_f, D_r),$ where updates are restricted to the selected parameter subset $\theta_s$.
\end{definition}

\begin{definition}[Memorization Score \cite{wei2025memorization}]
Memorization score for a data point $x_i$ with label $y_i$ in a training set $D_{train}$, $(x_i,y_i)\in D_{train}$ with a randomized training algorithm $\mathcal{A}$ is given as follows,
\begin{equation*}
    \text{mem}(\mathcal{A}, D, x_i)= \Pr_{f_\theta \sim \mathcal{A}(D_{train})}(f_\theta(x_i)=y_i)-\Pr_{f_{\theta'} \sim \mathcal{A}(D_{train} \setminus (x_i,y_i))}(f_{\theta'}(x_i)=y_i),
\end{equation*}
where $f_\theta$ is a model parameterized by $\theta$ sampled from the distribution $\mathcal{A}(D_{train})$, and $f_{\theta'}$ is a model parameterized by $\theta'$ sampled from the distribution $\mathcal{A}(D_{train})\setminus (x_i,y_i)$ . The first term considers model parameters trained on the entire dataset, while the second is the model parameters
trained without the example $(x_i,y_i)$. A high memorization score indicates that excluding the example
causes a significant change in the performance of the model for that example.
\end{definition}
\section{Proposed Similarity Aware Localized Unlearning}
We propose a Similarity-Aware Localized Unlearning framework that extends the Deletion by Example Localization (DEL) \cite{torkzadehmahani2024improved}, by explicitly incorporating intermediate feature-space similarity between forget and retain set images into the unlearning process. Existing localization-based unlearning methods \cite{torkzadehmahani2024improved} identify influential layers and neurons primarily from the forget set $(D_f)$, without considering whether these parameters also contribute to semantically similar samples in the retain set $(D_r)$. As a result, modifying localized parameters may affect learned representation from the retain set.

Our key observation is that semantically similar forget and retain samples are embedded in nearby regions of the intermediate feature space (in Fig. \ref{fig:intro_similar}) and therefore tend to activate overlapping subsets of neurons. Motivated by this observation, we perform localization using information from both the forget and retain sets. To address this limitation, our method uses forget samples and constrains the unlearning process with the retain set to preserve their representations. By jointly considering parameter localization and sample similarity, we selectively modify forget-specific knowledge while minimizing collateral damage to the $D_r$. By jointly analyzing neuron importance across $D_f$ and $D_r$, our proposed method identifies parameters that are strongly associated with the forget set while accounting for their contribution to similar retain samples. This similarity-aware localization enables targeted removal of forget-specific knowledge while reducing collateral damage to retained knowledge and preserving overall model performance.

\begin{definition}[Semantic Similarity]
Let $\phi(\cdot)$ denote the embedding representation produced by a model. For two samples $x_i$ and $x_j$, the semantic similarity between their learned representations is defined using cosine similarity as
\begin{equation}
\mathrm{sim}(x_i, x_j)
=
\frac{\phi(x_i)^\top \phi(x_j)}
{\|\phi(x_i)\| \, \|\phi(x_j)\|}.
\end{equation}\label{cossim}
\end{definition}
Two examples are considered semantically similar when their embedding representations exhibit high cosine similarity in the learned feature space. Consider a forget-set example $(x_i, y_i) \in D_f$ and a retain-set example $(x_j, y_j) \in D_r$, such that $\mathrm{sim}(x_i, x_j)$ is high. Since both examples occupy nearby regions in the learned representation space, they are likely to share intermediate feature representations and depend on overlapping subsets of model parameters. 

\begin{definition}[Similarity-Aware Localized Unlearning]
Consider $\theta^o = \{\theta_k\}_{k=1}^{N}$ denote the parameters of the original model $f_{\theta^o}$ and  $\text{mem}(\theta_k, x_i)$ quantify the contribution of parameter $\theta_k$ to the memorization of example $x_i$. An unlearning algorithm $\mathcal{U}$ is said to perform similarity-aware localized unlearning if it selects a subset of parameters $\theta_s \subset \theta^o, |\theta_s| = n < N,$ such that the selected parameters have high memorization contribution to the forget-set data while minimizing disruption to semantically similar retain-set data. Formally, the selected subset should satisfy the following,
\begin{equation}
    \text{mem}(\theta_s, x_i)
\gg
\text{mem}(\theta_s, x_j),
\end{equation}
for semantically similar pairs $(x_i, x_j)$ where $x_i \in D_f$ and $x_j \in D_r$.
\end{definition}
\subsection{Criticality scores}
\begin{definition}[(Parameter) criticality score]
    The \textit{(parameter) criticality score} $s_j$ of the $j^{\text{th}}$ parameter $\theta^o$ denoted as $\theta_j^o$, where $j \in \{1, \dots, N\}$ and $N$ is the total number of parameters, is defined as
\begin{equation}
\label{critscore}
    s_j = \left| \theta_j^o \cdot g(\theta_j^o, D) \right|.
\end{equation}
\end{definition}
The (parameter) criticality score measures the importance of the parameter $\theta_j^o$ with respect to a given dataset $D$. Let $g(\theta_j^o, D)$ denote the gradient of the loss with respect to $\theta_j^o$ computed over the forget set. This identifies parameters whose current values are both large and strongly implicated in the model’s predictions on the dataset. Each layer of the model consists of a set of $M$ neurons. We aim to identify the neurons with the highest (parameter) criticality scores; to that end, we introduce the following definition. 
\begin{definition}[Neuron criticality score]
    Let $\tilde{s}_i$ denote the sequence of (parameter) criticality scores corresponding to the parameters associated with neuron $o_i$, sorted in descending order. We compute the \textit{neuron criticality score} $co_i$ for neuron $o_i$, by averaging the top-$h$ parameter criticality scores:
\begin{equation}
    co_i = \frac{1}{h} \sum_{j=1}^{h} \tilde{s}_i[j].
\end{equation}
\end{definition}
Without loss of generality, in the remainder of this work we will refer to the (parameter) criticality scores simply as criticality scores.

\subsection{Criticality score of similarity aware localized unlearning}
We compute two sets of criticality scores for our localized unlearning framework. The first set is computed on the forget set $D_f$, denoted as $s^f$, and the second set is computed on the retain set $D_r$ denoted as $s^r$. The neuron criticality scores are then used to identify the neurons that should be finetuned for unlearning. 
We propose three methods of localization strategies by considering the forget set and retain set: (1) the \texttt{difference} method, (2) the \texttt{weighted\_difference} method, and (3) the \texttt{ratio} method. The methods are outlined below.

\subsubsection{Difference method}
In the \texttt{difference} method, we used the difference between the forget set and the retain set criticality score of a neuron to identify its relevance and later for unlearning. So, the criticality score is:
\begin{equation}
    \tilde{s}_j = \max\!\left(s_j^f - s_j^r,\; 0\right),
\end{equation}

where $s_j^f$ and $s_j^r$ are the forget-set and retain-set parameter criticality scores, respectively. The clamp at zero ensures that parameters more important for the retain set than the forget set receive a score of zero and are excluded from the localization mask entirely.

\subsubsection{Weighted Difference method}
In the \texttt{weighted\_difference} method, we introduce a scalar weight $w$ on the forget set criticality score as follows:
\begin{equation}
    \tilde{s}_j = \max\!\left(w \cdot s_j^f - s_j^r,\; 0\right).
\end{equation}
The scalar weight $w$ allows the balance between forget-set specificity and retain-set protection to be tuned continuously. As $w$ increases, the scoring criterion converges toward the forget-only baseline; as $w$ approaches 1, the scoring criterion converges to the unweighted difference.

\subsubsection{Ratio method} 
In the \texttt{ratio} method, we propose a ratio between the criticality score of the forget set and retain set:
\begin{equation}
    \tilde{s}_j = \frac{s_j^f}{s_j^r + \varepsilon},
\end{equation}
where $\varepsilon = 10^{-8}$ ensures numerical stability, favoring forget-specific parameters.

\subsection{Masking with criticality score}
A binary localization mask $m \in \{0, 1\}^p$ over all $p$ million trainable parameters is constructed by ranking neurons across all layers, in descending order of criticality score based on our three proposed methods, and greedily selecting neurons until the cumulative parameter count of selected channels meets a budget $\alpha = 30\%$ of total parameters, similarly to \cite{torkzadehmahani2024improved}. The budget is applied globally, thus the mask is not constrained to act uniformly across layers, so that layers with higher aggregate criticality naturally contribute more selected channels.

\subsection{Reinitialization based on criticality score and unlearning}
\label{sec:reset}
After constructing the mask, we re-initialise each selected weight $\theta_j$ using Kaiming normal initialization \cite{he2015delving}, $2 / n_{\text{in}}(j)$, where $n_{\text{in}}(j)$ is the number of input connections to the layer containing $\theta_j$ \cite{he2015delving}. This matches the initialization used to train the original model and preserves activation variance during the forward pass; unselected weights remain unchanged during re-initialization. The model is then fine-tuned on the retain set $D_r$ for the same number of epochs as in \cite{torkzadehmahani2024improved}. During fine-tuning, gradients of unselected weights are masked, restricting updates to the re-initialised parameter subset.

\section{Experiments}
\subsection{Dataset}
 
 

We use the CIFAR-10 dataset \cite{krizhevsky2009learning}, which contains 60,000 colour images (\(32 \times 32\)) from 10 classes, split into 50,000 training and 10,000 test images. The training set is further divided into 40,000 training and 10,000 validation images, where the validation set is used only for model selection. For the unlearning task, the 40,000 training images are partitioned into a forget set \(D_f\) of 5,000 images and a retain set \(D_r\) of 35,000 images using a fixed random seed. The forget set comprises 12.5\% of the training data and is sampled uniformly at random, preserving the class distribution of the original dataset.
\subsubsection{Retain-similar evaluation dataset:}
A key component of the evaluation strategy is the construction of a \textit{retain-similar set} $D_{r,\text{sim}}$: a subset of the retain set consisting of the retain-set examples most semantically similar to the forget-set examples under the original model's learned representations. Semantic similarity is measured using cosine similarity between 512-dimensional embeddings from the original model $f_{\theta^o}$. Embeddings are computed in a single forward pass for all 5,000 forget-set and 35,000 retain-set samples. For each forget-set sample $x_f \in D_f$, the nearest neighbour is selected from retain-set samples $x_r \in D_r$ of the same class ($y_f = y_r$):
\begin{equation}
    x_r^* = \underset{x_r \in \mathcal{D}_r : y_r = y_f}{\arg\max}\; \text{sim}(x_f, x_r),
\end{equation}
where $\text{sim}(x_f, x_r)$ is computed by \eqref{cossim}. The retain-similar set $D_{r,\text{sim}}$ is formed by collecting the unique nearest neighbours $\{x_r^*\}_{x_r^* \in D_r}$, deduplicated, so that each retain example appears at most once. In practice, the resulting set contains a slightly smaller number of examples than the forget set, since multiple forget-set examples share a nearest neighbour. We used 4,575 samples from $D_{r,\text{sim}}$.
\subsection{Model architecture}
We use a modified ResNet-18 \cite{he2016deep} for all experiments. Because the standard ImageNet configuration aggressively downsamples CIFAR-10 images ($32\times32$), we replace the initial $7 \times 7$ stride-2 convolution with a $3 \times 3$ stride-1 convolution and remove the first max-pooling layer. 
We have two models for experiments and for evaluating our proposed method,
\begin{itemize}
    \item \textbf{Original Model} - The original model $f_{\theta^o}$ was trained on the full training set using SGD \cite{bottou2012stochastic} with a learning rate of 0.1, momentum of 0.9, weight decay of $5\times10^{-4}$, batch size of 128, and 50 epochs. The learning rate followed a cosine annealing schedule \cite{loshchilov2017sgdr} with a minimum of 0.001 similar to \cite{torkzadehmahani2024improved}. The model achieved 89.87\% test accuracy. As expected for CIFAR-10, automobile (96.20\%) and ship (94.90\%) had the highest accuracy, whereas cat (78.30\%), dog (84.10\%), and bird (85.50\%) had the lowest \cite{krizhevsky2009learning}.
    \item \textbf{Oracle Model} - The oracle model $f_{\theta^r}$ is trained with all the data except the forget set $D_f$. We use the hyperparameter similar to the original model for training except the number of epochs as 35. The oracle model achieved a test set accuracy of 85.91\%, a retain set accuracy of 100\%, and a forget set accuracy of 86.32\%. 
\end{itemize}
\subsection{Evaluation measures}
Unlearning quality is evaluated against an oracle model retrained from scratch on $D_r$, which serves as the gold standard \cite{bourtoule2020machineunlearning}. We report six metrics: test accuracy, retain accuracy, forget accuracy, retain-similar accuracy, membership inference attack (MIA) score \cite{shokri2017membership}, and KL divergence \cite{golatkar2020eternal}. Metrics are reported for the original model $f_{\theta^o}$, the unlearned model $f_{\theta^u}$, and the oracle model $f_{\theta^r}$. Except for KL divergence, performance is summarized by the delta as follows,
\begin{equation}
\Delta_{\text{metric}} = \text{metric}(f_{\theta^r}) - \text{metric}(f_{\theta^u}),
\end{equation}
where values closer to zero indicate better agreement with the oracle. Positive values indicate that the unlearned model underperforms the oracle, while negative values indicate over-correction. The retain-similar accuracy delta measures collateral damage to semantically similar retained examples.

\textbf{Test accuracy} measures generalisation to the held-out test set:
\begin{equation}
    \Delta_{\text{test}} = \text{acc}_{\text{test}}(f_{\theta^r}) - \text{acc}_{\text{test}}(f_{\theta^u})
\end{equation}
A large positive delta indicates collateral damage to the model's general representations.

\textbf{Retain accuracy} measures performance on the full retain set $\mathcal{D}_r$:
\begin{equation}
    \Delta_{\text{retain}} = \text{acc}_{\text{retain}}(f_{\theta^r}) - \text{acc}_{\text{retain}}(f_{\theta^u})
\end{equation}
A significant positive delta indicates that the unlearning procedure has perturbed parameters important for examples that should be preserved.

\textbf{Forget accuracy} measures whether the model has ceased to correctly classify the forgotten examples:
\begin{equation}
    \Delta_{\text{forget}} = \text{acc}_{\text{forget}}(f_{\theta^r}) - \text{acc}_{\text{forget}}(f_{\theta^u})
\end{equation}
Following \cite{torkzadehmahani2024improved}, the target is a forget-set accuracy matching the oracle's: too high indicates insufficient forgetting; too low indicates over-forgetting, itself a detectable deviation from the oracle.

\textbf{Retain-similar accuracy} measures performance on $D_{r,\text{sim}}$:
\begin{equation}
    \Delta_{\text{rs}} = \text{acc}_{r,\text{sim}}(f_{\theta^r}) - \text{acc}_{r,\text{sim}}(f_{\theta^u})
\end{equation}
This is the primary metric for the central hypothesis of our proposal. If retain-aware criticality scoring is effective, $\Delta_{\text{rs}}$ should be smaller than the corresponding delta produced by the forget-only baseline, indicating that the more conservative channel selection has reduced collateral damage to the retained examples most semantically proximate to the forgotten data.

\textbf{KL divergence} provides a distributional measure of the similarity between the unlearned model and the oracle on the forget set \cite{golatkar2020eternal,nguyen2025survey}:
\begin{equation}
    D_{\text{KL}}\!\left(p_{f_{\theta^u}} \;\|\; p_{f_{\theta^r}}\right)
    = \frac{1}{|
    D_f|} \sum_{x \in D_f} \sum_{y \in Y}
    p_{f_{\theta^u}}(y \mid x) \log \frac{p_{f_{\theta^u}}(y \mid x)}{p_{f_{\theta^r}}(y \mid x)}
\end{equation}
where $p_{f_{\theta^u}}(y \mid x)$ and $p_{f_{\theta^r}}(y \mid x)$ are the softmax output distributions of the unlearned and oracle models respectively. A value of zero indicates that the two models are behaviourally indistinguishable on $D_f$. Unlike accuracy-based metrics, KL divergence captures differences in the full output distribution rather than just the predicted class, providing sensitivity to subtle residual memorization that accuracy-based metrics may miss \cite{golatkar2020eternal}.


\textbf{MIA score} measures forgetting quality using the confidence-based membership inference attack of \cite{fan2024salunempoweringmachineunlearning}. Following their protocol, an RBF-kernel SVC \cite{broomhead1988radial,cortes1995support} is trained to distinguish retain-set samples (members) from test-set samples (non-members) using the maximum softmax probability as the attack feature. The trained classifier is then applied to the forget set, whose examples are treated as non-members. The MIA score is defined as follows,
\begin{equation}
\text{MIA}_{\text{score}} =
\frac{|\{x_f \in D_f : \hat{y}(x_f)=\text{non-member}\}|}{|D_f|}.
\end{equation}
\subsection{Results and Discussion}
Table \ref{maintab} summarizes the delta metrics for the baseline \cite{torkzadehmahani2024improved} and our proposed method. We subsample data from retain set $D_r$ to compute the criticality score for the retain set. 
\begin{table}[H]
\centering
\small
\begin{tabular}{l l c r r r r r r l}
\hline
\textbf{Method} & \textbf{Sub} & $w$ & $\Delta_{\text{test}}$ & $\Delta_{\text{retain}}$ & $\Delta_{\text{forget}}$ & $\Delta_{\text{rs}}$ & $\Delta_{\text{MIA}}$ & \text{KL} \\
\hline
\texttt{forget\_only(baseline)}   & --    & --  & +2.58 & +0.49 & +2.70 & +0.31 & +5.48 & 0.443 \\
\hline
\textbf{\texttt{difference}}     & \textbf{Yes}   & \textbf{--}  & \textbf{+0.64} & +0.11 & \textbf{+0.56} & +0.04 & +4.98 & 0.395 \\
\texttt{difference}     & No    & --  & +1.16 & +0.20 & +1.24 & +0.11 & +4.16 & \textbf{0.380}\\
\hline
\texttt{w\_diff}        & Yes   & 3   & +1.49 & +0.45 & +1.42 & +0.35 & +4.30 & 0.423\\
\texttt{w\_diff}        & No    & 3   & +1.58 & +0.40 & +1.96 & +0.28 & +5.90 & 0.424\\
\texttt{w\_diff}        & Yes   & 5   & +2.32 & +0.35 & +1.96 & +0.28 & +5.52 & 0.432\\
\texttt{w\_diff}        & No    & 5   & +1.67 & +0.44 & +1.84 & +0.33 & +6.22 & 0.429\\
\texttt{w\_diff}        & Yes   & 7   & +2.42 & +0.55 & +1.90 & +0.33 & +5.80 & 0.413\\
\texttt{w\_diff}        & No    & 7   & +2.48 & +0.48 & +2.90 & +0.44 & +4.58 & 0.438\\
\hline
\texttt{ratio}          & Yes   & --  & $-$1.16 & \textbf{+0.01} & $-$3.86 & \textbf{+0.02} & $-$0.34 & 0.479\\
\texttt{ratio}          & No    & --  & $-$1.34 & \textbf{+0.02} & $-$3.74 & \textbf{+0.00} & \textbf{+0.56} & 0.482\\
\hline
\end{tabular}
\caption{Unlearning results across baseline and our proposed method: Delta metrics for all eleven experimental conditions. \textbf{Sub} denotes subsampled retain set used for criticality computation, $\Delta_{\text{rs}}$ denotes retain-similar accuracy delta. Positive values indicate the unlearned model performs worse than the oracle, negative values indicate over-correction beyond the oracle, and zero indicates perfect oracle-matching. \textbf{Bold letters} indicate the best results.}
\label{maintab}
\end{table}A random subsample (indicated as column header \textbf{Sub}) of  5{,}000 examples drawn from $D_r$, while non-subsampled used all 35{,}000 data in $D_r$. For our weighted difference method, we choose three values: $w \in \{3, 5, 7\}$, each in both subsampled and non-subsampled  variants. A value of $w = 1$ imposes  the strictest retain-set protection; large $w$ reduces to the forget-only  baseline. The three values tested span a range that is expected to capture  the transition between these behaviours: $w = 3$ imposes moderate retain-set protection, $w = 7$ approaches the forget-only regime, and $w = 5$ falls between them.
\subsubsection{Discussion of forget-only baseline}
The \texttt{forget\_only} baseline has $\Delta_{\text{forget}}=+2.70$, indicating mild over-forgetting relative to the oracle, and $\Delta_{\text{test}}=+2.58$, showing utility loss. Its  $\Delta_{\text{rs}}=+0.49$ supports the feature-entanglement hypothesis: forget-only localization selects channels also used by visually similar retain examples, causing collateral damage on $D_{r,\text{sim}}$. The large $\Delta_{\text{MIA}}=+5.48$ and KL divergence of $0.395$ further show stronger residual membership signal and poorer oracle alignment than retain-aware methods. Overall, \texttt{forget\_only} damages retain performance while failing to match the oracle distributionally.
\subsubsection{Discussion of similarity aware localized unlearning}
\paragraph{\textbf{Difference method}} The subsampled \texttt{difference} method produces the notable result among our proposed approaches. It achieves a $\Delta_{\text{forget}}$ of +0.56, a $\Delta_{\text{test}}$ of +0.64, a $\Delta_{\text{retain}}$ +0.11, and critically $\Delta_{\text{rs}}$ of +0.04 — the smallest among all conditions that achieves confirms  true forgetting. The $\Delta_{\text{MIA}}$ of +4.98 indicating that the unlearned model most closely resembles the oracle from the adversary's perspective. The KL divergence of 0.395 is the second lowest among valid conditions, confirming that the subsampled difference method produces the output distribution on the forget set that is very close to the oracle's — a result that complements and extends the accuracy-based findings.

The non-subsampled difference method remains valid, but its $\Delta_{\text{rs}}=+0.11$ is higher than the subsampled variant. This may reflect the retain set’s long-tail structure \cite{feldman2020does}: full retain aggregation includes atypical examples whose channel usage overlaps with the forget set, shifting the mask away from the forget-similar boundary region. In contrast, subsampling emphasizes denser head-region examples, yielding retain scores better aligned with $D_{r,\text{sim}}$ and therefore lower collateral damage.
\paragraph{\textbf{Weighted difference method}} The weighted-difference variants are valid across all values of $w$, but larger $w$ mainly makes the intervention more aggressive. In the subsampled setting, $\Delta_{\text{test}}$ increases from $+1.49$ to $+2.42$, while $\Delta_{\text{MIA}}$ increases from $+4.30$ to $+5.80$, suggesting greater utility loss and residual membership signal as forget-set weighting increases. Subsampling usually improves $\Delta_{\text{forget}}$, except at $w=5$, but its effect on $D_{r,\text{sim}}$ is mixed: non-sampling is better at $w=3$, while subsampling is better at $w=5$ and $w=7$. The KL values remain close overall, indicating similar forget-set distributional alignment. Overall, subsampled $w=3$ gives the best balance across $\Delta_{\text{test}}$, $\Delta_{\text{forget}}$, and $\Delta_{\text{MIA}}$.
\paragraph{\textbf{Ratio method}} The ratio conditions preserve utility but fail to forget. Both variants have strongly negative $\Delta_{\text{forget}}$ values, $-3.86$ and $-3.74$, showing substantial under-forgetting relative to the oracle. Meanwhile, $\Delta_{\text{test}}$, $\Delta_{\text{retain}}$, and $\Delta_{\text{rs}}$ remain close to zero or negative, indicating little collateral damage. This is not a favourable trade-off, since the retained utility comes from an insufficiently effective intervention. The likely cause is the ratio criterion $s^f_j/(s^r_j+\epsilon)$, which can over-prioritise channels with small retain scores even when their absolute forget importance is weak, while downweighting entangled channels that are important to both forget and retain examples. The similar results for subsampled and non-subsampled variants suggest that the failure is intrinsic to the ratio formulation rather than the retain-set size. The elevated KL values further indicate poor oracle alignment on the forget set.
\begin{figure}[htbp]
\centering
\includegraphics[width=\textwidth]{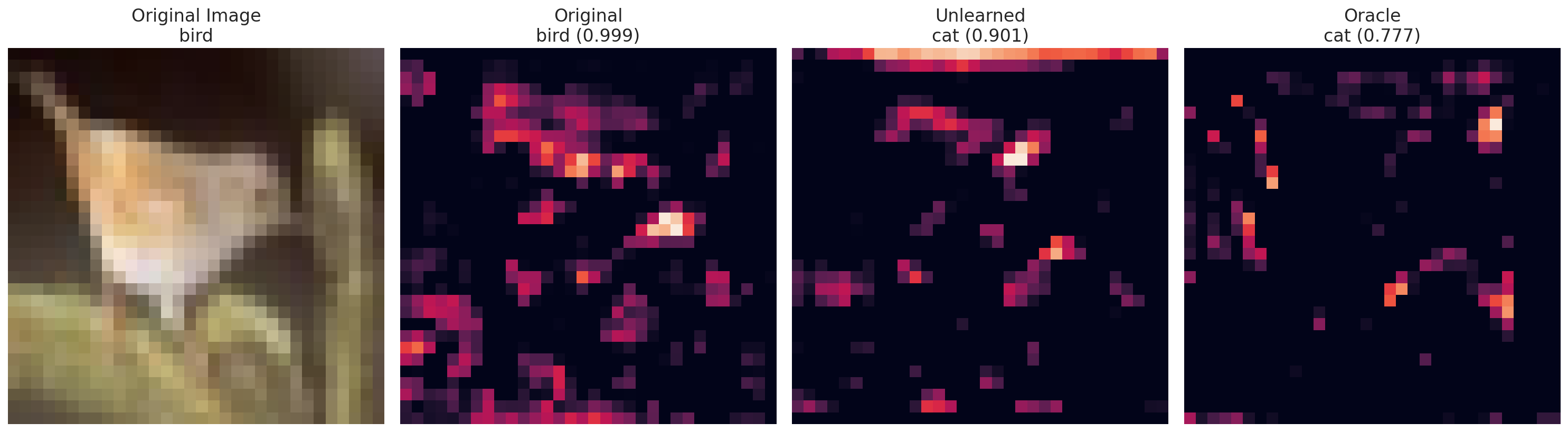}
\caption[Subsampled difference method: forget-set bird example (misclassification).]{Forget-set bird. Original (conf.\ 0.999): correctly classified with sparse scattered activations. Unlearned (conf.\ 0.901): misclassifies as cat; activation partially smeared, indicating loss of bird-discriminative features. Oracle (conf.\ 0.777): also misclassifies as cat, confirming the image sits near the bird/cat decision boundary.}
\label{fig:gradcam_diff_sub_forget7}
\end{figure}
\vspace{-0.5cm}
\begin{figure}[htbp]
\centering
\includegraphics[width=\textwidth]{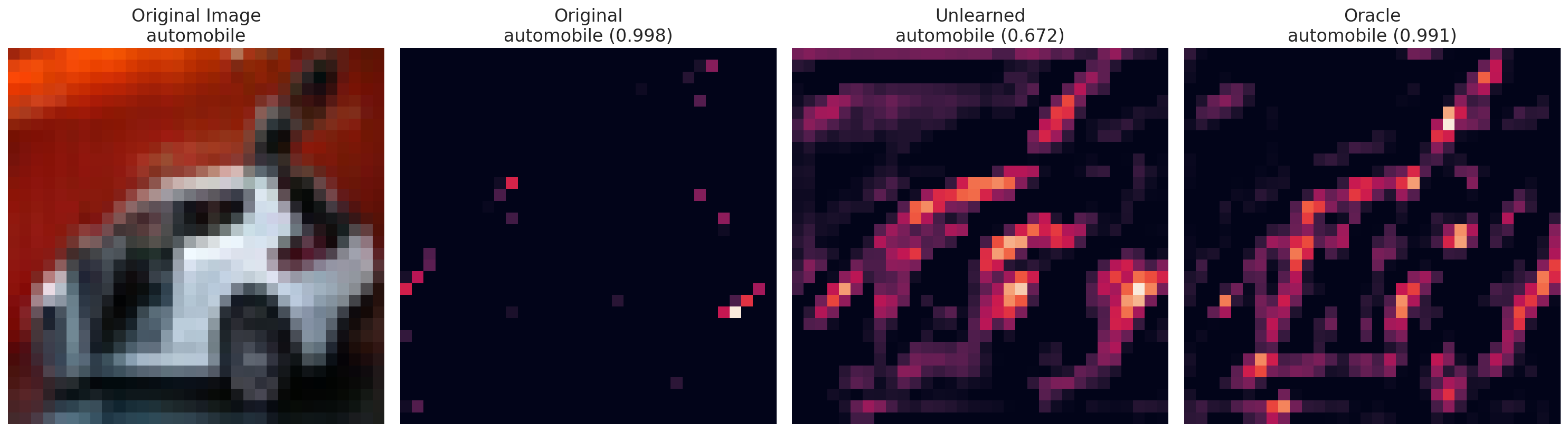}
\caption[Subsampled difference method: test-set automobile example.]{Test-set automobile. Original (conf.\ 0.998): near-zero \texttt{layer1} activation despite high confidence, indicating classification driven by deeper layers. Unlearned (conf.\ 0.672): strong structured activation along car edges and body lines. Oracle (conf.\ 0.991): similarly structured edge-following activation, closely resembling the unlearned model.}
\label{fig:gradcam_diff_sub_test6}
\end{figure}
\begin{figure}[htbp]
\centering
\includegraphics[width=\textwidth]{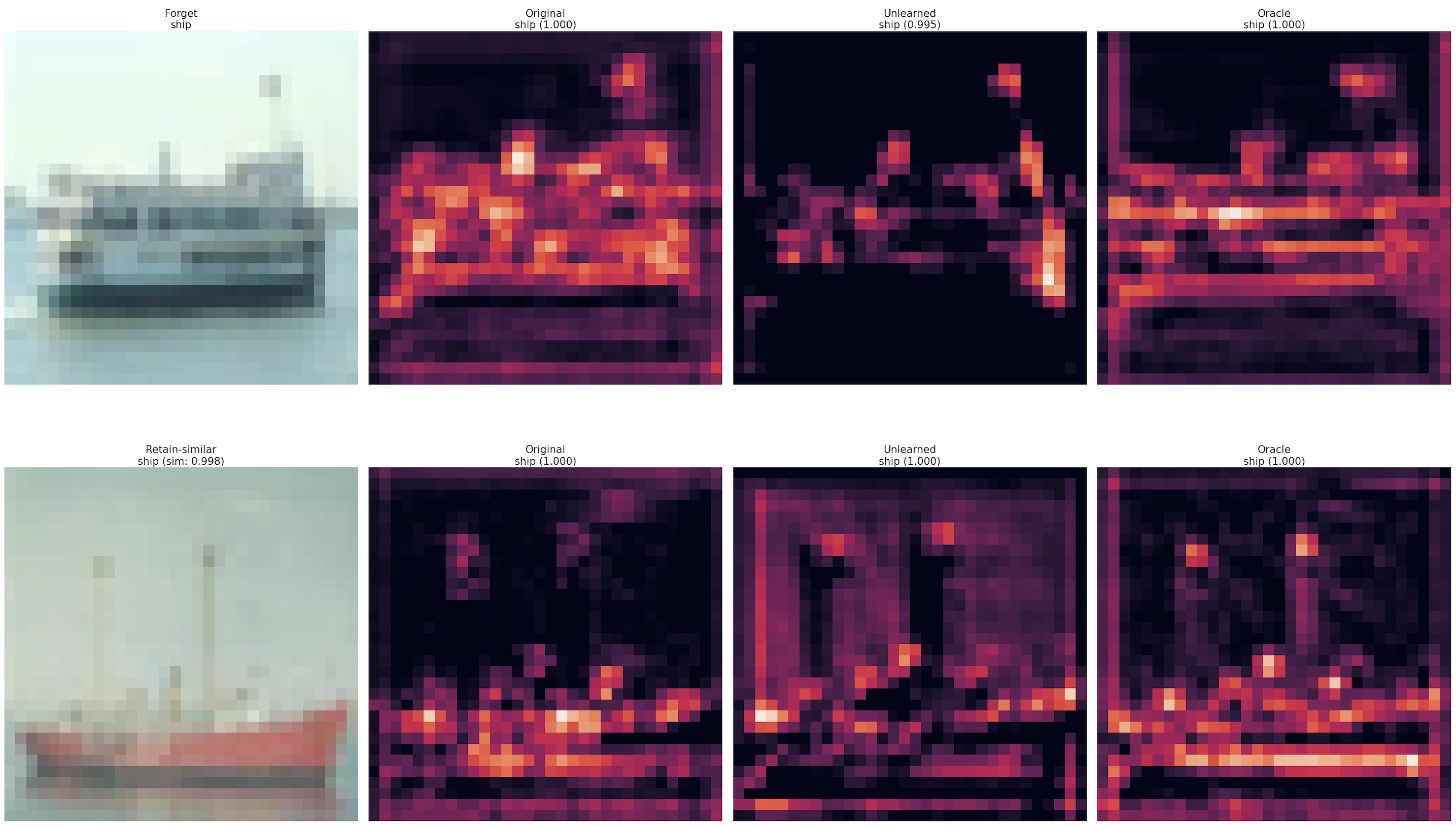}
\caption[Retain-similar ship pair (semantic similarity 0.998).]{Forget/retain-similar ship pair (cosine similarity 0.998). Top: unlearning changes the forget-set activation from broad to sparse and localized. Bottom: the retain-similar image remains correctly classified with ship-focused attention, indicating limited collateral damage.}
\label{fig:gradcam_rank39}
\end{figure}
\subsubsection{Discussion of $\Delta_{\text{rs}}$ metric}
Across conditions, $\Delta_{\text{rs}}$ is generally smaller in magnitude than $\Delta_{\text{retain}}$. This can be explained by the construction of $D_{r,\text{sim}}$, which selects retain examples closest to the forget set in embedding space. In our setting, these examples are likely to lie in relatively dense regions with many similar training neighbours, whereas the full retain set also contains rarer, more atypical long-tail examples. Prior work by Feldman et al. \cite{feldman2020does,feldman2020neural} shows that memorization is especially important for such rare examples, while high-frequency examples can often be recovered from shared, generalizable features. Consequently, after a partial reset and fine-tuning, examples in $D_{r,\text{sim}}$ may be easier to recover than long-tail retain examples, leading to smaller $\Delta_{\text{rs}}$ than $\Delta_{\text{retain}}$. The key point is therefore not that forget-only localization causes large absolute damage to $D_{r,\text{sim}}$, but that it causes more collateral damage than retain-aware methods. The purpose of $\Delta_{\text{rs}}$ is to measure this excess damage on the retain examples most similar to the forgotten data.
\subsection{GradCAM-based Visual analysis}
We use Grad-CAM \cite{selvaraju2017grad} to compare the original, unlearned, and oracle models. We analyze three examples in total: one each from the forget set, test set, and retain-similar set.
\paragraph{Forget-set examples}
The forget-set bird in Fig. \ref{fig:gradcam_diff_sub_forget7} is correctly classified by the original model, which shows broad, scattered activations. Both the unlearned and oracle models predict cat, while the unlearned activation is more diffuse, suggesting disruption of bird-specific features. Its agreement with the oracle indicates oracle-aligned forgetting rather than an unlearning failure.
\paragraph{Test-set examples.} The test-set automobile in Fig. \ref{fig:gradcam_diff_sub_test6} is correctly classified by the original model with high confidence (0.998), despite weak Grad-CAM activation at the selected layer. After unlearning, activation becomes more structured along the vehicle’s edges and contours, closely resembling the oracle and suggesting recovery of an oracle-like feature representation during retain-set fine-tuning.
\paragraph{Retain-similar examples.} Fig. \ref{fig:gradcam_rank39} shows a forget-set example and its nearest retain-set neighbour (cosine similarity 0.998). For the forget-set image, the original model exhibits broad activations across the hull and surrounding structure, whereas the unlearned model produces sparser, more localized activations with lower confidence, even than the oracle model. This suggests that unlearning has altered the original representation and reduced reliance on features previously used for classification. For the retain-similar image, the unlearned model still attends to the ship hull and classifies it correctly, indicating that the retained representation is largely preserved. However, its more diffuse activations indicate only partial preservation of the original feature representation.
\paragraph{Computational time}
Experiments were run on NVIDIA A100 and L40 GPUs, with runtimes reported in MM:SS. Training the original and oracle models took 07:46 and 04:46, respectively. The forget-only baseline required 02:57 in total (00:02 for criticality computation and 02:54 for fine-tuning), while the difference method required 02:59 (00:02 and 02:57), indicating negligible overhead.

\section{Conclusion}
We introduced a similarity-aware localized unlearning framework that considers parameter importance for both forgotten and retained data. A retain-similar evaluation set was also proposed to measure collateral damage directly. Experiments on CIFAR-10 with ResNet-18 showed that the subsampled difference method achieved the best balance between effective forgetting and retained utility. These results demonstrate that retain-aware localization preserves shared representations more effectively than forget-only localization, with negligible computational overhead. Future work should evaluate the method on larger datasets, architectures, and unlearning settings.
\section*{Acknowledgments}
M. Kumaran was supported by The University of Manchester and the Manchester Unit of the European Laboratory for Learning and Intelligent Systems (ELLIS). The research of V. Kouni was supported in part by the French National Research Agency under the France 2030 program, reference ANR-23-PEIA-0003. H. Harikumar was supported by the UKRI Turing AI World-Leading Researcher Fellowship (EP/W002973/1), UKRI AI Hub in Generative Models (EP/Y028805/1), European Lighthouse of AI for Sustainability (ELIAS, 10080425), and Laboratory for AI Security Research (LASR). The views expressed in this paper are those of the authors and do not necessarily reflect the position of LASR or His Majesty's Government. The authors would like to acknowledge the assistance by Research IT and the use of the Computational Shared Facility at The University of Manchester.  

%

\bibliographystyle{splncs04}
\bibliography{refs}
\newpage
\section*{Appendix}
\begin{figure}[htbp]
\centering
\includegraphics[width=\textwidth]{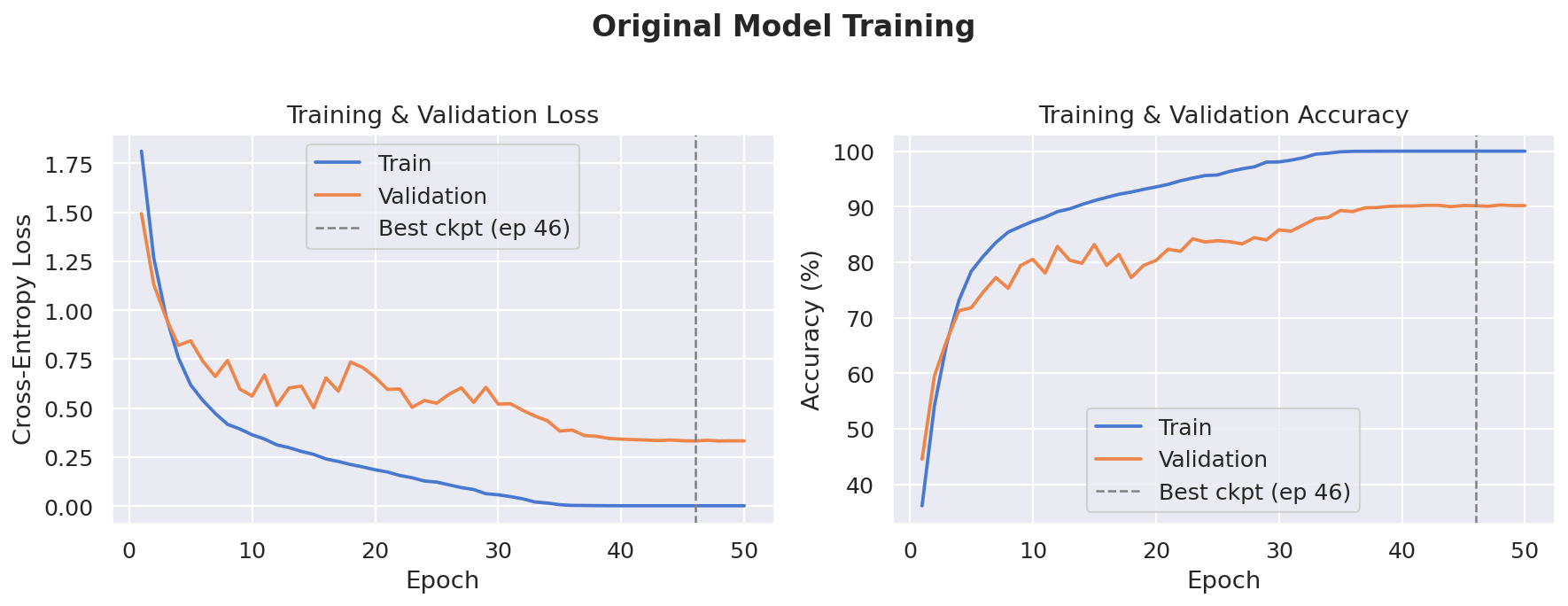}
\caption[Original model training curves.]{Training and validation loss (left) and accuracy (right) for the original model $\mathcal{M}(D)$ over 50 epochs. Training loss decreases monotonically to near-zero while validation loss stabilises around 0.33, reflecting the growing train-validation gap characteristic of overfitting to training examples including $\mathcal{D}_f$. Validation accuracy plateaus around 90\% from epoch 35 onward with minor fluctuations. The dashed line marks the best checkpoint (model) at epoch 46.}
\label{fig:original_training_curves}
\end{figure}
Fig. ~\ref{fig:original_training_curves} shows the training and validation curves for the original model. The original model $\mathcal{M}(D)$ was trained for 50 epochs in the entire 40 {,} 000 sample training subset and achieved a test set accuracy of 89.87\%, consistent with expected performance for a well-tuned ResNet-18 in CIFAR-10 \cite{he2016deep}. The training trajectory followed the expected pattern: rapid initial learning in the first ten epochs, a slower refinement phase as the cosine learning rate schedule decayed, and training accuracy saturating near 100\% by epoch 39 while validation accuracy continued to improve gradually. The best model was saved at epoch 46. The original model achieves 100\% accuracy on both the retain and forget sets, as expected for a fully converged model evaluated on its own training data.
\begin{table}[htbp]
\centering
\begin{tabular}{l c c c c c}
\hline
\textbf{Class} & \text{Original (\%)} & \text{Unlearned (\%)} & \text{Oracle (\%)} & \text{$\Delta_{\text{org,unlearn}}$ (pp)} & \text{$\Delta_{\text{oracle,unlearn}}$ (pp)} \\
\hline
Airplane    & 90.90 & 87.00 & 88.50 & +3.90 & +1.50 \\
Automobile  & 96.20 & 94.10 & 93.60 & +2.10 & $-$0.50 \\
Bird        & 85.50 & 77.80 & 79.60 & +7.70 & +1.80 \\
Cat         & 78.30 & 68.90 & 67.40 & +9.40 & $-$1.50 \\
Deer        & 89.30 & 84.10 & 87.10 & +5.20 & +3.00 \\
Dog         & 84.10 & 80.10 & 77.90 & +4.00 & $-$2.20 \\
Frog        & 93.10 & 89.10 & 90.90 & +4.00 & +1.80 \\
Horse       & 93.20 & 89.00 & 89.30 & +4.20 & +0.30 \\
Ship        & 94.90 & 93.40 & 93.00 & +1.50 & $-$0.40 \\
Truck       & 93.20 & 89.20 & 91.80 & +4.00 & +2.60 \\
\hline
Overall & 89.87 & 85.27 & 85.91 & 4.60 & 0.64 \\
\hline
\end{tabular}
\vspace{0.2cm}
\caption[Per-class test accuracy: original model vs subsampled difference unlearned model vs oracle.]{Per-class test set accuracy for the original model, the subsampled \texttt{difference} unlearned model, and the oracle model. $\Delta_{\text{org,unlearn}}$ (pp) denotes the difference in percentage points between the original and unlearned model. $\Delta_{\text{oracle,unlearn}}$ (pp) denotes the difference in percentage points between the oracle and unlearned model. The difference is largest for cat and bird between the unlearned model and original model; the unlearned model closely tracks the oracle, suggesting effective unlearning.}
\label{tab:per_class_comparison}
\end{table}
Table~\ref{tab:per_class_comparison} shows the per-class test accuracy for the original model, oracle model, and our unlearned model (based on subsampled \texttt{difference}). The largest drops are in cat ($9.4$ pp) and bird ($7.7$ pp) — the two most visually similar and hardest CIFAR-10 classes \cite{krizhevsky2009learning} — suggesting the unlearning intervention disproportionately disrupts discriminative features near class boundaries. Visually distinctive classes such as automobile ($2.1$ pp) and ship ($1.5$ pp) suffer the smallest drops, consistent with their features being more separable and therefore less affected by the channel reset.
\begin{table}[H]
\centering
\small
\begin{tabular}{l l c r r r}
\hline
\textbf{Method} & \textbf{Sub} & \textbf{w} &\textbf{Crit time} & \textbf{Finetune time} & \textbf{Total time} \\
\hline
Original model         & --  & -- & --    & --    & 07:46 \\
Oracle model           & --  & --  & --    & --    & 04:46 \\
\hline
Baseline (del)         & --  & -- & 00:02 & 02:54 & 02:57 \\
\hline
\texttt{Ours (difference)} & Yes & -- & 00:02 & 02:57 & 02:59 \\
\texttt{Ours (difference)} & No  & -- & 00:06 & 02:57 & 03:03 \\
\hline
\texttt{Ours (weighted)}   & Yes & 3  & 00:02 & 02:54 & 02:56 \\
\texttt{Ours (weighted)}   & No  & 3  & 00:06 & 02:55 & 03:01 \\
\texttt{Ours (weighted)}   & Yes & 5  & 00:02 & 02:53 & 02:55 \\
\texttt{Ours (weighted)}   & No  & 5  & 00:06 & 02:55 & 03:01 \\
\texttt{Ours (weighted)}   & Yes & 7  & 00:02 & 02:57 & 02:59 \\
\texttt{Ours (weighted)}   & No  & 7  & 00:06 & 02:58 & 03:04 \\
\hline
\texttt{Ours (ratio)}      & Yes & -- & 00:02 & 02:58 & 03:00 \\
\texttt{Ours (ratio)}      & No  & -- & 00:05 & 02:56 & 03:02 \\
\hline
\end{tabular}
\vspace{0.2cm}
\caption[Wall-clock time per stage across all eleven conditions (seed 1).]{Wall-clock time (MM:SS, minutes and seconds, rounded to the nearest second) taken per pipeline stage for all eleven experimental conditions. Sub denotes whether a subsampled retain set was used for criticality computation. Original is the from-scratch training of the base model; Oracle is the retain-only retraining used as the gold-standard reference; Crit is the criticality-scoring step alone; and Finetune is the subsequent finetuning stage. Original and Oracle have no unlearning pipeline, so only total training time is shown.}
\label{tab:time}
\end{table}
Table~\ref{tab:time} reports wall-clock times for all experimental conditions. The criticality-scoring step is negligible in all cases, taking at most six seconds even without subsampling, meaning the total pipeline cost is dominated almost entirely by finetuning. All unlearning methods complete in under three minutes and five seconds, compared to seven minutes and forty-six seconds for original training and four minutes and forty-six seconds for oracle retraining. Subsampling reduces criticality computation time from roughly five--six seconds to two seconds across all variants, with no meaningful effect on finetuning time, making it a cost-free efficiency gain. 
\begin{table}[H]
\centering
\small
\begin{tabular}{l l c r r r r r r}
\hline
\textbf{Method} & \textbf{Sub} & \textbf{Finetuning Size} & $\Delta_{\text{test}}$ & $\Delta_{\text{retain}}$ & $\Delta_{\text{forget}}$ & $\Delta_{\text{rs}}$ & $\Delta_{\text{MIA}}$ & \text{KL} \\
\hline
\texttt{forget\_only} & -- & 500   & +65.11 & +78.76 & +64.80 & +77.79 & +11.12 & 1.812 \\
\texttt{forget\_only} & -- & 1000  & +48.75 & +62.45 & +49.60 & +61.09 & +10.98 & 1.484 \\
\texttt{forget\_only} & -- & 2000  & +35.25 & +47.04 & +34.16 & +43.98 & $-$12.74 & 1.089 \\
\texttt{forget\_only} & -- & 5000  & +22.79 & +33.73 & +21.60 & +30.67 & $-$30.18 & 0.802 \\
\texttt{forget\_only} & -- & 10000 & +13.96 & +19.94 & +13.24 & +16.90 & $-$11.06 & 0.625 \\
\texttt{forget\_only} & -- & 20000 & +5.91  & +8.48  & +6.72  & +7.32  & +0.88  & 0.479 \\
\texttt{forget\_only} & -- & 35000 & +2.58  & +0.49  & +2.70  & +0.31  & +5.48  & 0.443 \\
\hline
\texttt{difference}   & Yes & 500   & +54.48 & +68.19 & +55.54 & +65.84 & $-$38.50 & 1.692 \\
\texttt{difference}   & Yes & 1000  & +48.52 & +61.40 & +48.82 & +59.02 & $-$60.10 & 1.572 \\
\texttt{difference}   & Yes & 2000  & +37.06 & +49.12 & +37.28 & +45.68 & $-$47.32 & 1.316 \\
\texttt{difference}   & Yes & 5000  & +26.62 & +34.83 & +26.32 & +31.06 & $-$32.72 & 0.995 \\
\texttt{difference}   & Yes & 10000 & +13.49 & +19.54 & +14.22 & +15.91 & $-$14.64 & 0.643 \\
\texttt{difference}   & Yes & 20000 & +4.68  & +7.78  & +5.28  & +5.99  & $-$0.62  & 0.441 \\
\texttt{difference}   & Yes & 35000 & +0.64  & +0.11  & +0.56  & +0.04  & +4.98  & 0.395 \\
\hline
\end{tabular}
\vspace{0.2cm}
\caption{Effect of finetuning retain set size on unlearning quality. Sub denotes subsampled retain set used for criticality computation, Finetuning size denotes the size of the retain set used for finetuning}
\label{tab:finetuning_size}
\end{table}
Table~\ref{tab:finetuning_size} examines how the size of the retain set used during finetuning affects unlearning quality, comparing the \texttt{forget\_only} baseline against the subsampled \texttt{difference} method. Both methods exhibit a consistent trend: smaller finetuning sets yield larger deviations from the oracle across all metrics, with deltas decreasing monotonically as retain set size grows toward 35{,}000. At 35{,}000 samples, the \texttt{difference} method achieves the smallest overall deltas, with $\Delta_{\text{test}} = 0.64$, $\Delta_{\text{forget}} = 0.56$, and $\Delta_{\text{rs}} = 0.04$, closely tracking the oracle on all retain-side metrics. These results indicate that finetuning set size is a key determinant of unlearning quality, and that the full retain set of 35{,}000 samples is necessary to achieve oracle-level performance.
\begin{table}[htbp]
\centering
\small
\begin{tabular}{l c c c}
\hline
\textbf{Layer} & \textbf{Total channels} & \textbf{Selected} & \textbf{Fraction (\%)} \\
\hline
\texttt{conv1}                     & 64  & 31  & 48.4 \\
\texttt{layer1.0.conv1}            & 64  & 40  & 62.5 \\
\texttt{layer1.0.conv2}            & 64  & 46  & 71.9 \\
\texttt{layer1.1.conv1}            & 64  & 27  & 42.2 \\
\texttt{layer1.1.conv2}            & 64  & 49  & 76.6 \\
\texttt{layer2.0.conv1}            & 128 & 94  & 73.4 \\
\texttt{layer2.0.conv2}            & 128 & 101 & 78.9 \\
\texttt{layer2.0.downsample.0}     & 128 & 106 & 82.8 \\
\texttt{layer2.1.conv1}            & 128 & 71  & 55.5 \\
\texttt{layer2.1.conv2}            & 128 & 78  & 60.9 \\
\texttt{layer3.0.conv1}            & 256 & 228 & 89.1 \\
\texttt{layer3.0.conv2}            & 256 & 226 & 88.3 \\
\texttt{layer3.0.downsample.0}     & 256 & 229 & 89.5 \\
\texttt{layer3.1.conv1}            & 256 & 115 & 44.9 \\
\texttt{layer3.1.conv2}            & 256 & 97  & 37.9 \\
\texttt{layer4.0.conv1}            & 512 & 305 & 59.6 \\
\texttt{layer4.0.conv2}            & 512 & 116 & 22.7 \\
\texttt{layer4.0.downsample.0}     & 512 & 371 & 72.5 \\
\texttt{layer4.1.conv1}            & 512 & 58  & 11.3 \\
\texttt{layer4.1.conv2}            & 512 & 0   & 0.0  \\
\texttt{fc}                        & 10  & 10  & 100.0 \\
\hline
\textbf{Total parameters}          & \multicolumn{3}{c}{3{,}352{,}873 / 11{,}173{,}962 (30.01\%)} \\
\hline
\end{tabular}
\vspace{0.2cm}
\caption{Channels selected per layer under the subsampled \texttt{difference} method with $\alpha = 30\%$ parameter budget. Selection is highest in layer3 (up to 89.5\%) where forget-set criticality is concentrated, and unevenly distributed in layer4: \texttt{layer4.0.downsample.0} has 72.5\% of channels selected while \texttt{layer4.1.conv2} has none, reflecting the differential criterion zeroing out channels whose retain-set importance meets or exceeds their forget-set importance. The fully connected layer is selected in its entirety, indicating that all output logit weights are implicated in the forget-set predictions.}
\label{tab:mask_channels}
\end{table}
Table~\ref{tab:mask_channels} shows the channel selection produced by our subsampled \texttt{difference} method, which is used as the representative condition. The selection is markedly non-uniform across layers: layer3 has the majority of its channels selected (up to 229/256 in \texttt{layer3.0.downsample.0}), reflecting the criticality scoring concentrating forget-set importance in the deeper, more class-specific residual stages. By contrast, \texttt{layer4.1.conv2} contributes zero selected channels despite its depth — a consequence of the differential criterion: its parameters are sufficiently important to the retain set that the difference score is zeroed out entirely. Earlier layers show moderate, more evenly distributed selection, consistent with their role in encoding lower-level features shared across both forget and retain examples.
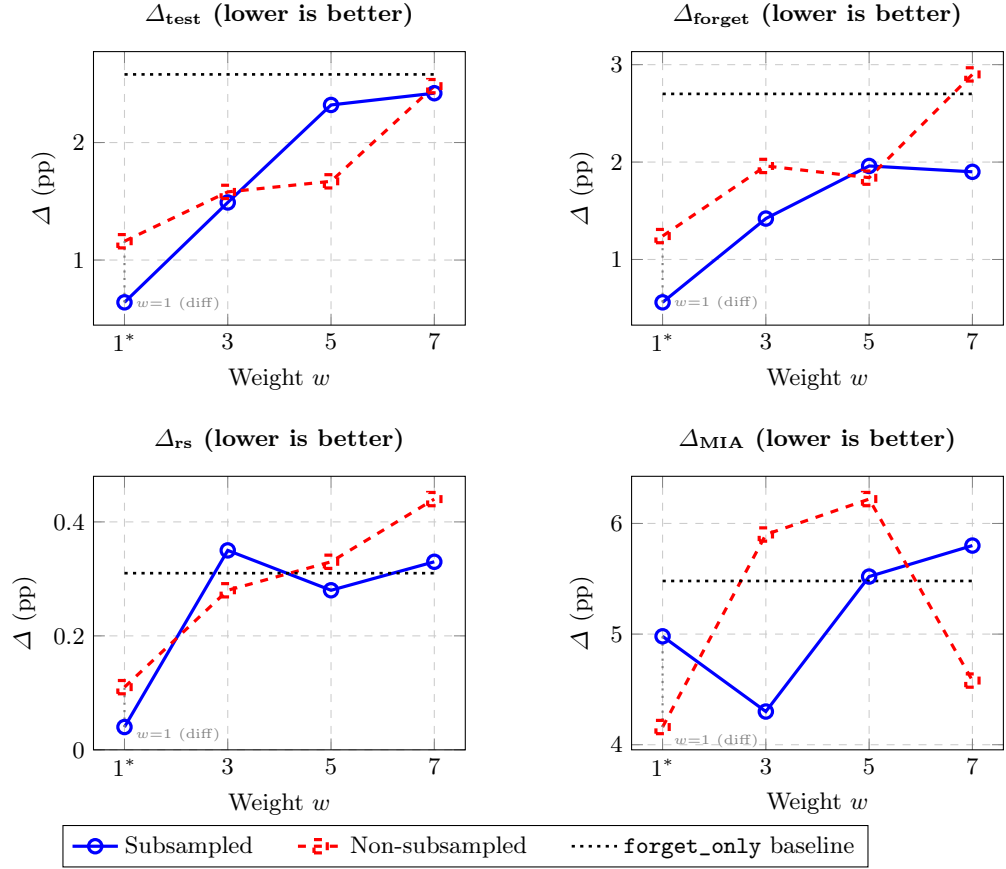
\begin{figure}[htbp]
\centering
\begin{tikzpicture}
\begin{groupplot}[
    group style={
        group size=2 by 2,
        horizontal sep=2.2cm,
        vertical sep=2.0cm,
    },
    width=6.5cm,
    height=5.2cm,
    xlabel={Weight $w$},
    xtick={1,3,5,7},
    xticklabels={$1^*$,3,5,7},
    grid=major,
    grid style={dashed, gray!40},
    tick label style={font=\small},
    label style={font=\small},
    title style={font=\small\bfseries},
    every axis plot/.append style={line width=1.2pt, mark size=2.5pt},
]
\nextgroupplot[title={$\Delta_{\text{test}}$ (lower is better)}, ylabel={$\Delta$ (pp)}]
\addplot[color=blue, mark=o]
    coordinates {(1,0.64) (3,1.49) (5,2.32) (7,2.42)};
\addplot[color=red, mark=square, dashed]
    coordinates {(1,1.16) (3,1.58) (5,1.67) (7,2.48)};
\addplot[color=black, dotted, line width=1.0pt]
    coordinates {(1,2.58) (7,2.58)};
\draw[dotted, thick, gray] (axis cs:1,0.64) -- (axis cs:1,1.16);
\node[font=\tiny, gray, anchor=south west] at (axis cs:1.05,0.50) {$w{=}1$ (diff)};
\node[font=\tiny, black, anchor=south east] at (axis cs:6.9,2.58) {};
\nextgroupplot[title={$\Delta_{\text{forget}}$ (lower is better)}, ylabel={$\Delta$ (pp)}]
\addplot[color=blue, mark=o]
    coordinates {(1,0.56) (3,1.42) (5,1.96) (7,1.90)};
\addplot[color=red, mark=square, dashed]
    coordinates {(1,1.24) (3,1.96) (5,1.84) (7,2.90)};
\addplot[color=black, dotted, line width=1.0pt]
    coordinates {(1,2.70) (7,2.70)};
\draw[dotted, thick, gray] (axis cs:1,0.56) -- (axis cs:1,1.24);
\node[font=\tiny, gray, anchor=south west] at (axis cs:1.05,0.40) {$w{=}1$ (diff)};
\node[font=\tiny, black, anchor=south east] at (axis cs:6.9,2.70) {};
\nextgroupplot[title={$\Delta_{\text{rs}}$ (lower is better)}, ylabel={$\Delta$ (pp)}]
\addplot[color=blue, mark=o]
    coordinates {(1,0.04) (3,0.35) (5,0.28) (7,0.33)};
\addplot[color=red, mark=square, dashed]
    coordinates {(1,0.11) (3,0.28) (5,0.33) (7,0.44)};
\addplot[color=black, dotted, line width=1.0pt]
    coordinates {(1,0.31) (7,0.31)};
\draw[dotted, thick, gray] (axis cs:1,0.04) -- (axis cs:1,0.11);
\node[font=\tiny, gray, anchor=south west] at (axis cs:1.05,0.00) {$w{=}1$ (diff)};
\node[font=\tiny, black, anchor=south east] at (axis cs:6.9,0.31) {};
\nextgroupplot[
    title={$\Delta_{\text{MIA}}$ (lower is better)},
    ylabel={$\Delta$ (pp)},
    legend to name=grouplegend,
    legend style={
        legend columns=3,
        font=\small,
        /tikz/every even column/.append style={column sep=0.5cm},
    },
]
\addplot[color=blue, mark=o]
    coordinates {(1,4.98) (3,4.30) (5,5.52) (7,5.80)};
\addlegendentry{Subsampled}
\addplot[color=red, mark=square, dashed]
    coordinates {(1,4.16) (3,5.90) (5,6.22) (7,4.58)};
\addlegendentry{Non-subsampled}
\addplot[color=black, dotted, line width=1.0pt]
    coordinates {(1,5.48) (7,5.48)};
\addlegendentry{\texttt{forget\_only} baseline}
\draw[dotted, thick, gray] (axis cs:1,4.98) -- (axis cs:1,4.16);
\node[font=\tiny, gray, anchor=south west] at (axis cs:1.05,3.90) {$w{=}1$ (diff)};
\node[font=\tiny, black, anchor=south east] at (axis cs:6.9,5.48) {};
\end{groupplot}
\end{tikzpicture}
\vspace{0.5cm}
\centering\ref{grouplegend}
\caption[Metric trends as a function of weight $w$.]{Metric deltas as a function of weight $w$ for subsampled and non-subsampled conditions. The \texttt{difference} method is treated as $w=1$ (marked $1^*$ on the $x$-axis). The black dotted horizontal line marks the \texttt{forget\_only} baseline score on each metric as a reference. Each metric is reported as $\Delta = \text{oracle} - \text{unlearned}$; lower values indicate closer oracle-matching. The dotted vertical segment at $w=1$ highlights the gap between the two subsampling conditions at the unweighted case.}
\label{fig:weight_trends}
\end{figure}
Fig. ~\ref{fig:weight_trends} visualises how each metric varies with $w$ across both subsampled and non-subsampled conditions, treating the unweighted \texttt{difference} method as $w=1$. The figure reveals several consistent patterns. On the retain-similar metric, the subsampled \texttt{difference} method at $w=1$ achieves the lowest delta of all conditions, and performance degrades as $w$ increases toward the baseline, confirming that stronger forget-set weighting erodes retain-set protection. The non-subsampled variant consistently underperforms its subsampled counterpart on $\Delta_{\text{rs}}$. The test accuracy panel shows a similar ordering at $w=1$, with the subsampled \texttt{difference} method achieving the best scores before degrading as $w$ increases.
\end{document}